\documentclass{article} %
\usepackage[preprint]{colm2025_conference}

\usepackage{microtype}
\usepackage{hyperref}
\usepackage{url}
\usepackage{booktabs}
\usepackage{enumitem}

\usepackage{lineno}

\definecolor{darkblue}{rgb}{0, 0, 0.5}
\hypersetup{colorlinks=true, citecolor=darkblue, linkcolor=darkblue, urlcolor=darkblue}

\usepackage{graphicx}
\usepackage{wrapfig}

\usepackage{tcolorbox}
\usepackage{algorithm}
\usepackage{algpseudocode}

\usepackage{subcaption}
\usepackage{multirow}
\usepackage{amsmath, amssymb}

\newcommand{\method}{Ours}

\title{Efficient Model Selection for Time Series Forecasting via LLMs}

\author{Wang Wei\\
Department of Computer Science\\
Virginia Tech\\
Blacksburg, VA, USA \\
\texttt{wangwei718@vt.edu} \\
\And
Tiankai Yang \\
Department of Computer Science \\
University of South California \\
Los Angeles, CA, USA\\
\texttt{tiankaiy@usc.edu} \\
\And
Hongjie Chen \\
Dolby Labs\\
Atlanta, GA, USA \\
\texttt{hongjie.chen@dolby.com}
\And 
Ryan A. Rossi\\
Adobe Research\\
San Jose, CA, USA\\
\texttt{ryrossi@adobe.com}
\And
Yue Zhao\\
Department of Computer Science \\
University of South California \\
Los Angeles, CA, USA\\
\texttt{yzhao010@usc.edu} \\
\And
Franck Dernoncourt\\
Adobe Research\\
Seattle, WA, USA\\
\texttt{dernonco@adobe.com}
\And
Hoda Eldardiry\\
Department of Computer Science\\
Virginia Tech\\
Blacksburg, VA, USA \\
\texttt{hdardiry@vt.edu}
}

\begin{document}

\ifcolmsubmission
\linenumbers
\fi

\maketitle

\begin{abstract}
Model selection is a critical step in time series forecasting, traditionally requiring extensive performance evaluations across various datasets. Meta-learning approaches aim to automate this process, but they typically depend on pre-constructed performance matrix, which is costly to build. In this work, we propose to leverage Large Language Models (LLMs) as a lightweight alternative for model selection. Our method eliminates the need for explicit performance matrix by utilizing the inherent knowledge and reasoning capabilities of LLMs. Through extensive experiments with Llama, GPT, and Gemini, we demonstrate that our approach outperforms traditional meta-learning techniques and heuristic baselines, while significantly reducing computational overhead. These findings underscore the potential of LLMs in efficient model selection for time series forecasting.
\end{abstract}

\section{Introduction} 
Time series forecasting plays a crucial role in a wide range of real-world applications, enabling informed decision-making and strategic planning across various domains, including finance \citep{sezer2019financial}, healthcare \citep{healthcare}, software monitoring \citep{sun2023efficient}, energy \citep{energy}, retail \citep{retail}, and weather prediction \citep{weather}. Selecting an appropriate forecasting model is often a labor-intensive process requiring domain expertise and extensive computational resources. 
Traditional time series forecasting methods typically require substantial domain expertise and manual effort in model design, feature engineering, and hyperparameter tuning. This challenge is further intensified by the findings of \citet{autoforecast}, which indicate that no single learning strategy consistently outperforms others across all forecasting tasks, due to the inherent diversity of time series data. Consequently, traditional methods often fail to deliver high-quality predictions across diverse application domains.
A straightforward but na\"ive solution would be evaluating the performance for thousands of models on a given dataset to identify the most suitable one. However, such an approach is impractical due to the excessive computational cost and training time required for model evaluation on each new dataset.

To address the impracticality of exhaustively evaluating all models for each new dataset, meta-learning has recently gained great popularity in applications demanding model selection such as anomaly detection and classification \citep{outlierdetection}, graph learning \citep{graphlearning}, and recommendation \citep{recommendation}, especially for forecasting \citep{autoforecast}, which could quickly infer the best forecasting model after training on the models’ performances on historical datasets and the time-series meta-features of these datasets. Even though \citet{autoforecast} selects the best performing forecasting algorithm and its associated hyper-parameters with a 42× median inference time reduction averaged across all datasets compared to the na\"ive approach, nearly all state-of-the-art meta-learning approaches still require the construction of a large performance matrix, consisting of evaluations of hundreds or even thousands of models across a vast collection of forecasting datasets. This performance matrix, while crucial for traditional meta-learning-based model selection, is extremely costly to obtain in practice. Each dataset-model pair must be exhaustively evaluated, which demands significant computational resources and time. Furthermore, this matrix is typically used in conjunction with a carefully engineered meta-feature vector extracted from each time-series dataset to train a meta-learning model that can generalize and infer the best model for new forecasting tasks. 

LLMs have demonstrated exceptional generalization and reasoning capabilities, positioning them as promising tools for automating model selection in time series forecasting. By leveraging zero-shot prompting techniques, LLMs can generate structured reasoning paths without the need for task-specific exemplars. For instance, \citet{llmzeroshot} introduced a method where appending the phrase "Let's think step by step" to a prompt enables LLMs to perform complex reasoning tasks effectively. Building upon this, \citet{enhancingcot} proposed the Zero-shot Uncertainty-based Selection (ZEUS) approach, which enhances chain-of-thought (CoT) prompting by utilizing uncertainty estimates to select effective demonstrations without requiring access to model parameters. These advancements suggest that LLMs, through zero-shot and CoT prompting, can be harnessed to streamline model selection processes, reducing the need for exhaustive evaluations and manual interventions.

In this work, we propose an alternative paradigm: using LLMs to perform model selection without the need for an explicit performance matrix. Following the benchmark data specified in \citet{autoforecast}'s work, we investigate the effectiveness of LLMs in model selection for time series forecasting. Extensive experiments on over 320 datasets show that our method outperforms strategies such as directly selecting popular methods and even different meta-learning approaches\citep{isac} (including
simple and optimization-based meta-learners where a performance matrix is built and used during training). 

\textbf{Summary of Main Contributions.} The key contributions of this work are as follows:
\begin{itemize}[nosep, itemsep=3pt, leftmargin=*]
    \item \textbf{LLM-Driven Zero-Shot Model Selection for Time-Series Forecasting.} To the best of our knowledge, this work is the first to investigate the use of LLMs for selecting the most suitable time series forecasting model via zero-shot prompting. By evaluating multiple LLMs with various prompt designs, we demonstrate that LLM-based selection consistently outperforms both popular forecasting models and meta-learning approaches.
    \item \textbf{Computational Efficiency in Training and Inference.} Unlike conventional model selection techniques that require training and evaluation of multiple forecasting models and the costly pre-computed performance matrix required in traditional meta-learning, our approach leverages LLMs to infer the optimal model and hyperparameters instantly. This results in a significant reduction in computational overhead, making the method highly scalable and efficient for real-world forecasting applications.
    \item \textbf{Ablation Study on Prompt Design for Model Selection.} We conduct an ablation study to analyze the impact of incorporating meta-features and CoT reasoning in prompts across different LLMs. The findings could offer insights into effective prompt design strategies, guiding future improvements in LLM-driven model selection for time-series forecasting.
\end{itemize}

\section{Related Work}
\textbf{Model Selection in Time Series Forecasting.}
Model selection in time series forecasting has evolved through various methodologies, encompassing traditional statistical approaches, meta-learning techniques, and the emerging LLMs.

Traditional methods often rely on statistical criteria to choose the most suitable forecasting model. For instance, the average rank method evaluates multiple models across different datasets, selecting the one with the lowest average rank based on performance metrics \citep{traditionalmodelselection}. While straightforward, these methods can be computationally intensive and may not generalize well across diverse time series data. To overcome these limitations, \citet{metaLEMKE} explored meta-learning strategies that utilize characteristics of time series data to predict the performance of various forecasting models, facilitating more efficient and accurate model selection. Similarly, \citet{metaPRUDENCI} investigated meta-learning techniques to rank and select time series models based on extracted meta-features, demonstrating improved forecasting accuracy. Recently, \citet{autoforecast} have also demonstrated that meta-learning can be used to infer the best model given dataset characteristics and model space without needing an exhaustive evaluation of all existing models on a new dataset. However, these approaches still require constructing performance matrix that capture the evaluation results of all models across all datasets, which is computationally expensive and time-consuming.

\textbf{LLMs for Time Series Forecasting.}
The integration of LLMs into time series forecasting has garnered significant attention, with recent studies exploring their potential for model selection and prediction tasks. \citet{jin2024timellmtimeseriesforecasting} introduced Time-LLM, a framework that reprograms LLMs for time series forecasting by aligning time series data with natural language inputs. \citet{gruver2024largelanguagemodelszeroshot} demonstrated that LLMs, such as GPT-3 and Llama-2, can perform zero-shot time series forecasting by encoding time series as sequences of numerical digits, framing forecasting as a next-token prediction task. \citet{tempoprompt} introduced an interpretable prompt-tuning-based generative transformer for time series representation learning. \citet{survey_llm_ts} provided a comprehensive survey on the application of LLMs in time series analysis, highlighting their potential to enhance forecasting performance across various domains. However, these studies differ from our approach as they employ LLMs directly as forecasting models for new datasets, whereas our work focuses on leveraging LLMs for model selection. Specifically, we demonstrate how LLMs can effectively identify the most suitable forecasting model to achieve optimal predictive performance.

\textbf{Prompting.} Prompting has emerged as the primary approach for tailoring language models to various downstream applications. Zero-shot prompting enables LLMs to perform tasks without specific examples by appending phrases like "Let's think step by step" to the prompt, effectively eliciting reasoning process~\citep{llmzeroshot}. CoT prompting further improves multi-step reasoning by incorporating intermediate reasoning steps into the prompt, leading to better performance on complex tasks \citep{cot}. Surveys on prompt design strategies provide comprehensive overviews of techniques such as manual design and optimization algorithms, emphasizing their impact on LLM performance across diverse tasks \citep{li-2023-practical}. These developments underscore the critical role of prompt engineering in fully leveraging LLMs for complex reasoning and decision-making tasks.

\begin{figure}
    \centering
    \includegraphics[width=1\linewidth]{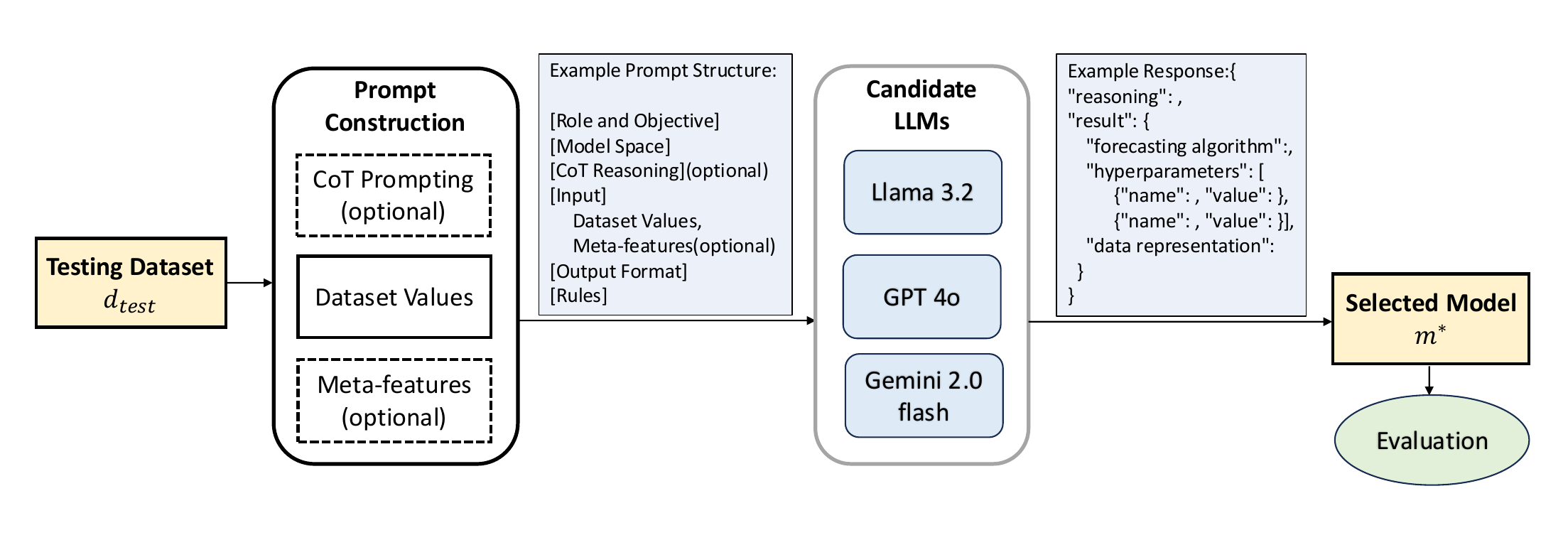}
    \caption{An overview of model selection via LLMs.}
    \label{fig:overview}
\end{figure}
  
\section{Methodology}

\subsection{Overview}
The model selection task for time series forecasting is formulated as a mapping from dataset-based prompts to candidate forecasting models. Let $S: \mathcal{P} \rightarrow \mathcal{M}$, where \(\mathcal{P}\) denotes the set of possible prompts and \(\mathcal{M}\) represents the space of candidate forecasting models. For each dataset \(d_i \in \mathcal{D}\), the process comprises three components:
\begin{itemize}[nosep, itemsep=3pt, leftmargin=*]
    \item \textbf{Prompt Construction}: Construct a prompt \(p_i \in \mathcal{P}\) from \(d_i\) using one of the predefined prompt templates. 
    \item \textbf{LLM-Based Model Selection}: The prompt \(p_i\) is submitted to an LLM to obtain the recommended model \(m_i = S(p_i)\), where \(m_i \in \mathcal{M}\).
    \item \textbf{Forecasting and Evaluation}: Apply \(m_i\) to \(d_i\) to produce forecasts and evaluate performance using appropriate metrics.
\end{itemize}

\begin{algorithm}[t]
\caption{Model Selection via LLMs}
\label{alg:llm_selection}
\begin{algorithmic}  

\State \textbf{Input:} Time series dataset \( d_{\text{test}} \), model space \( \mathcal{M} \), prompt template \( \mathcal{P} \)
\State \textbf{Output:} Selected forecasting model \( m^* \)

\vspace{0.5em}
\State \textbf{Step 1: Prompt Construction}
\State Get dataset values \( X_{\text{test}} \) from \( d_{\text{test}} \)
\If{Meta-features are included}
    \State Include meta-features \( F_{\text{test}} \) of \( d_{\text{test}} \)
\EndIf
\If{CoT is included}
    \State Incorporate reasoning steps into prompt
\EndIf
\State Generate prompt \( p \) using:
\[
p = \text{Format}(X_{\text{test}}, [F_{\text{test}}], [\text{CoT}])
\]

\vspace{0.5em}
\State \textbf{Step 2: Query LLM for Model Selection}
\State Obtain selected model \( m^* = (a^*, h^*, g^*) \) by querying the LLM:
\[
m^* = S(p)
\]
where \( a^* \) is the forecasting algorithm, \( h^* \) is the hyperparameter set, and \( g^* \) is the data representation.

\vspace{0.5em}
\State \textbf{Step 3: Forecasting and Evaluation}
\State Apply \( m^* \) to generate forecasts for \( d_{\text{test}} \)
\State Compute performance metrics

\vspace{0.5em}
\State \textbf{Return:} Selected model \( m^* \)

\end{algorithmic}
\end{algorithm}

\textbf{Problem Statement. } Given a new dataset \(d_{\text{test}}\) (i.e., unseen time series forecasting task), select a model \(m \in \mathcal{M}\) to employ on that dataset. 
\subsection{Prompt Construction}
We designed four distinct prompt structures, each varying in the inclusion of meta-features and CoT reasoning. The detailed structure of our prompts is illustrated in Appendix~\ref{appendix:a1}.
\begin{itemize}[nosep, itemsep=3pt, leftmargin=*]
    \item\textbf{Dataset Values Only}: Providing raw time series data.
    \item\textbf{Dataset Values and Meta-Features}: Combining raw data with pre-computed meta-features from \citet{autoforecast}. Details of meta-features are available ~\ref{meta_feats}.
    \item\textbf{Dataset Values with CoT}: Including raw data along with a step-by-step reasoning instruction in the prompt to guide the LLM.
    \item\textbf{Dataset Values and Meta-Features with CoT}: Integrating raw data, meta-features, and CoT reasoning.
\end{itemize}
\subsection{Model Selection}
The model space is denoted as:
\(\mathcal{M} = \{m_1, m_2, \dots\}\). Each model \(m_i \in \mathcal{M}\) is given by the tuple $m_i = (a_i, h_i, g_i(\cdot))$,
where \(a_i\) is the forecasting algorithm, \(h_i\) is the hyper-parameter vector associated with \(a_i\), and \(g_i(\cdot): \mathbb{R}^{n_i} \to \mathbb{R}^{n_i }\) is the time-series data representation (e.g., raw, exponential smoothing).

Unlike traditional meta-learning approaches that operate on a predefined, discrete model space, our method allows for an infinite and continuous model space, where hyperparameters \(h_i\) can take any real-valued configuration.

\subsection{Comparison with Meta-Learning}
Meta-learning methods typically rely on an extensive performance matrix:
\(
\mathbf{P} \in \mathbb{R}^{n \times m}
\)
where \(\mathbf{P}_{i,j}\) represents the performance of model \(M_j\) on dataset \(D_i\). This matrix is computationally expensive to construct and is essential for training meta-learners. In contrast, our approach eliminates the need for:
\begin{itemize}[nosep, itemsep=3pt, leftmargin=*]
    \item\textbf{Explicit performance matrix.} Our method does not require historical model-dataset performance mappings.
    \item\textbf{Feature engineering.} While meta-learners depend on carefully designed meta-features, our LLM-based selection can operate without them.

    \item\textbf{Fixed model spaces.} Our method does not restrict selection to a predefined set of models and hyperparameters.
\end{itemize}

\begin{table}[ht]
    \centering
    \resizebox{\columnwidth}{!}
    {
    \begin{tabular}{ccccc}
    \toprule
    \textbf{Methods}     &\textbf{hit@1 accuracy} {$\uparrow$}  & \textbf{hit@5 accuracy} {$\uparrow$} &\textbf{hit@10 accuracy} {$\uparrow$}  &\textbf{hit@50 accuracy} {$\uparrow$} \\
    \midrule
    Random selection     &0.31  &1.25  &3.63  &14.75 \\
    Popular Selection    &0  &1.33  &3.77  &19.94 \\
    $ISAC^{*}$  &0.82  &2.67  &4.10  &11.45 \\
    $MLP^{*}$  &0.62  &1.13  &4.51  &22.25 \\
    \midrule
    \multicolumn{1}{l}{\method-Llama3.2}      &  &  &  & \\
    \multicolumn{1}{r}{w. data}  &0.83  &3.84  &6.65  &26.27 \\
    \multicolumn{1}{r}{w. data+CoT} &0.62 &3.12 &5.82 &26.69\\
    \multicolumn{1}{r}{w. data+meta\_features} &\textbf{1.14}  &\textbf{4.47}  &\textbf{7.27}  &\textbf{29.60} \\
    \multicolumn{1}{r}{w. data+meta\_features+CoT} &1.14 &3.43 &6.44 &25.96\\
    \midrule
    \multicolumn{1}{l}{\method-GPT4o}     &  &  &  & \\
    \multicolumn{1}{r}{w. data}    &0.21  &2.39  &4.47  &21.39\\
    \multicolumn{1}{r}{w. data+CoT}    &0.10 &1.25 &4.36 &21.39\\
    \multicolumn{1}{r}{w. data+meta\_features}&0.62  &2.39  &4.88  &20.56 \\
    \multicolumn{1}{r}{w. data+meta\_features+CoT} &0.52 &2.60 &4.88 &21.91\\
    \midrule
    \multicolumn{1}{l}{\method-Gemini2.0 flash}     &  &  &  & \\
    \multicolumn{1}{r}{w. data} &0.21  &0.62  &3.53  &20.77 \\
    \multicolumn{1}{r}{w. data+CoT} &0.31&0.93&2.91&19.94\\
    \multicolumn{1}{r}{w. data+meta\_features} &0  &1.04  &2.80  &20.87 \\
    \multicolumn{1}{r}{w. data+meta\_features+CoT} &0.21 &1.35 &3.01 &17.13\\
    \bottomrule

    \end{tabular}}
    \caption{hit@$k$ Accuracy (the higher ({$\uparrow$}), the better) comparison of LLMs against the different baselines. $*$ denotes meta-learning methods which utilized performance matrix during training.
    }
    \label{tab:hitk}
\end{table}

\section{Experiments}
We evaluate our LLM-based model selection approach through a series of experiments designed to address the following research questions:  
\begin{enumerate}[nosep, itemsep=3pt, leftmargin=*]
    \item Does employing LLMs for time-series forecasting model selection improve performance compared to not using model selection or other techniques like meta-learners?
    \item How much reduction in inference time do LLM-based methods achieve over the naïve approach, and what is the associated token cost for model selection?
    \item To what extent do meta-features and CoT prompting contribute to model selection performance, computational efficiency, and token usage?  
\end{enumerate}
\subsection{Experiment Settings}

\subsubsection{Dataset and Metrics}

\textbf{Dataset Source.} We use the same dataset as \cite{autoforecast}, which consists of 321 forecasting datasets spanning various application domains, including finance, IoT, energy, and storage. These datasets include benchmark time series from Kaggle, Adobe real traces, and other open-source repositories.
For each dataset, we randomly sample time windows of fixed length ($=16$) to form our evaluation samples.

\textbf{Evaluation Metrics.} Our evaluation focuses on two primary metrics: hit@$k$ accuracy and average Mean Squared Error (MSE). Hit@$k$ accuracy quantifies whether the selected model ranks among the top \(k\) models based on ground truth performance, while MSE measures the forecast error magnitude. Formally, hit@k accuracy is defined as:  
\begin{equation}
    \text{hit@}k = \frac{1}{N} \sum_{i=1}^{N} \mathbb{I} \left( M^*_i \in \mathcal{M}_{\text{ranked}}^k (D_i) \right)
\end{equation}
where \(\mathcal{M}_{\text{ranked}}(D_i) \) denotes the set of models ranked by their performance for a given dataset \( D_i \), \( \mathbb{I}(\cdot) \) is an indicator function that equals 1 if \( M^*_i \) is within the top \( k \) models and 0 otherwise, and \( N \) is the total number of test datasets.

In addition, we record training and inference time, as well as token usage, to assess the computational efficiency and resource overhead of the approaches.

To make it fair to compare, we adopt the same model space as our baselines \(\mathcal{M}\) from \citet{autoforecast}, which comprises 322 unique models (see Table~\ref{modelspace} for the complete list). This model space pairs seven state-of-the-art time-series forecasting algorithms
with their corresponding hyperparameters and various data representation methods. In addition, we utilize the precomputed performance matrix from \citet{autoforecast} to evaluate our proposed methods.

\subsubsection{LLMs and Hardware}
To evaluate the effectiveness of different LLMs in the forecasting model selection task, we conducted experiments using three competitive models: Llama 3.2-3B-Instruct \citep{llama32}, GPT-4o \citep{gpt4o}, and Gemini 2.0 Flash \citep{gemini2flash}. In experiments with Llama 3.2-3B-Instruct, we utilized a single NVIDIA A100 GPU with 80GB of memory. GPT-4o and Gemini 2.0 Flash are accessed via API.

\subsection{Baselines}
We compare our proposed approach against various baseline methods. They fall into two categories: methods that do not perform explicit model selection and meta-learning-based approaches.

\textbf{No Model Selection.} In this category, the same fixed model configuration or an ensemble of all models is applied. We consider the following strategies:
\begin{enumerate}[nosep, itemsep=3pt, leftmargin=*]
    \item \textbf{Random Model.} A model configuration is randomly selected from the model space \(\mathcal{M}\) for each time-series dataset.
    \item \textbf{Popular Model.} The most widely used forecasting model, Prophet \citep{prophet}, is selected given its strong community support (e.g., over 19k stars on GitHub).
    \item \textbf{SOTA Model.} We consider seven state-of-the-art forecasting models. For each model, we create multiple configurations by adjusting hyperparameters and data representations, resulting in 10 to 72 variants per model, as detailed in Table~\ref{modelspace}. The variant that achieves the best average performance across all training datasets is selected.
\end{enumerate}

\textbf{Meta-learners.}
These approaches leverage performance matrix to guide model selection:
\begin{enumerate}[nosep, itemsep=3pt, leftmargin=*, start=4]
    \item \textbf{ISAC} \citep{isac}: This clustering-based method groups training datasets based on their extracted meta-features. For a new dataset, ISAC identifies the nearest cluster and selects the best-performing model within that cluster.
    \item \textbf{MLP.} Given the training datasets and selected time window, the MLP regressor directly maps the meta-features onto model performances by regression \citep{autoforecast}.
\end{enumerate}

\begin{table}[t]
    \centering
    \begin{tabular}{c|c|c}
    \toprule
    & \textbf{Methods} & \textbf{Mean Square Error} {$\downarrow$}\\
    \midrule
    \multirow{7}{8em}{No Model Selection}
        &Seasonal Na\"ive  &$0.0345 \pm 0.0382$ \\
        &DeepAR &$0.0164 \pm 0.0506$ \\
        &Deep Factors &$0.0217 \pm 0.0415$ \\
        &Random Forest &$0.0199 \pm 0.0398$  \\
        &Prophet &$0.0155 \pm 0.0295$ \\
        &Gaussian Process &$0.1661 \pm 0.2104$\\
        &VAR &$0.0602 \pm 0.1260$ \\
    \midrule
    \multirow{2}{8em}{Meta learner}
        &ISAC &\textbf{$\mathbf{0.0071 \pm 0.0145}$} \\
        &MLP & $0.0351 \pm 0.1186$\\
    \midrule
    \multirow{3}{8em}{LLM based}
        &Ours-Llama3.2  &$\underline{0.0081\pm0.0297}$ \\
        &Ours-GPT4o &$0.0234\pm0.0596$ \\
        &Ours-Gemini2.0 flash &$0.0169\pm0.0407$ \\
    \bottomrule
    \end{tabular}
    \caption{Results for one-step ahead forecasting (MSE; the lower ($\downarrow$) the better). The selected model by LLMs
yields second best performance compared to baseline meta-learners and SOTA methods.}
    \label{tab:mse}
\end{table}

\subsection{Overall Results}
\textbf{Superiority of LLM-based Methods in hit@$\mathbf{k}$ and MSE.}  
The results presented in Tables~\ref{tab:hitk} and~\ref{tab:mse} highlight the effectiveness of our LLM-based approach compared to all baseline methods. \textbf{Ours-LLaMA3.2} consistently outperforms other selection strategies across both hit@$k$ accuracy and mean squared error (MSE). For instance, Ours-LLaMA3.2 achieves 100.27\%, 92.83\%, 77.32\%, and 61.20\% higher hit@10 accuracy compared to Random Selection, Popular Selection, ISAC, and MLP, respectively. In terms of forecasting performance, the model selected by Ours-LLaMA3.2 achieves the second lowest MSE among all tested methods, outperforming traditional SOTA forecasting models while achieving performance comparable to the best meta-learning method. Notably, unlike meta-learning approaches that require an extensive precomputed performance matrix for training, our LLM-based method selects models instantly without training.

\begin{figure}
  \begin{subfigure}[t]{.45\textwidth}
    \centering
    \includegraphics[width=\linewidth]{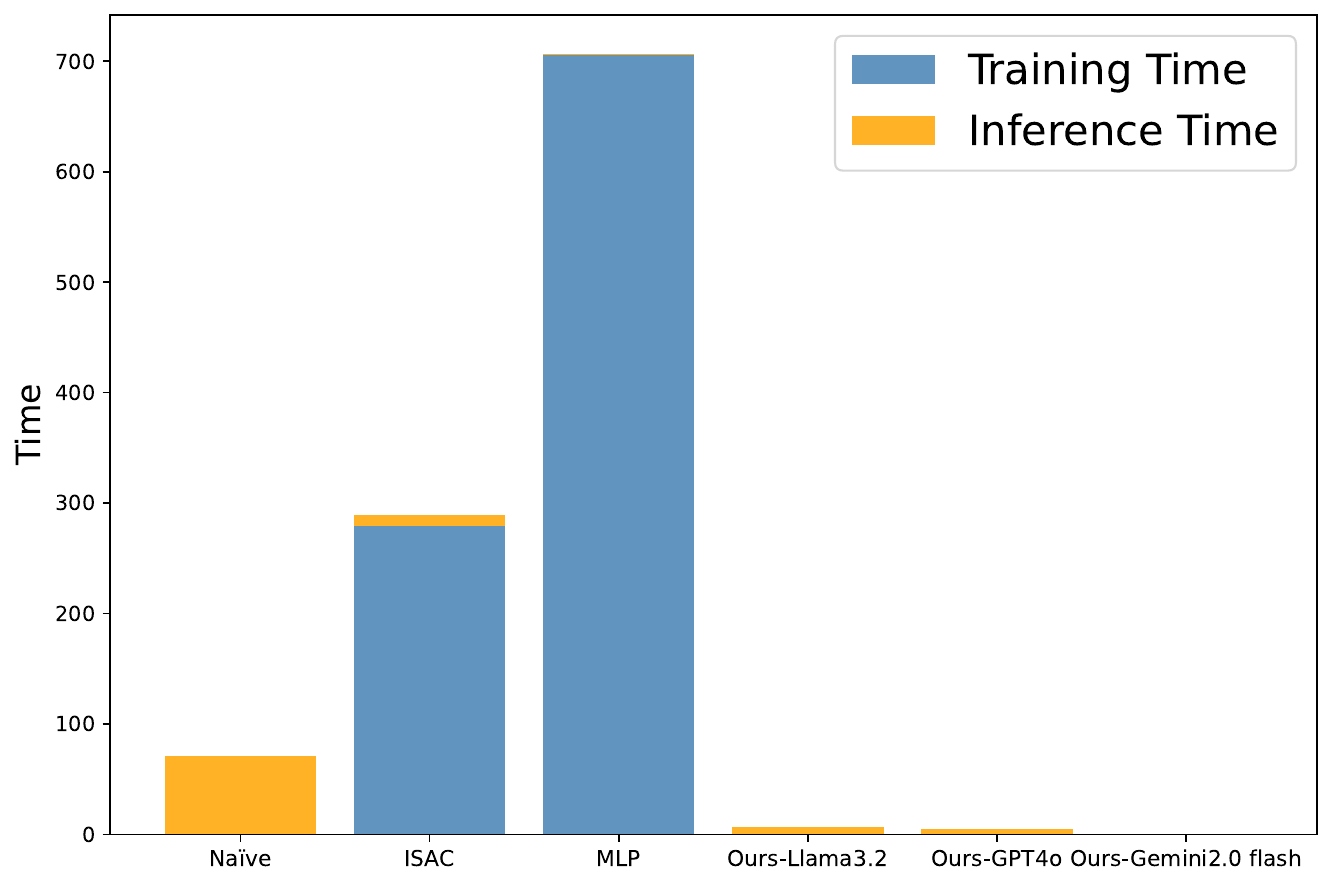}
    \caption{Average training and inference time (in seconds). Detailed mean and standard deviation values are provided in Table~\ref{time}.}
    \label{fig:time-avg}
  \end{subfigure}
  \hfill
  \begin{subfigure}[t]{.45\textwidth}
    \centering
    \includegraphics[width=\linewidth]{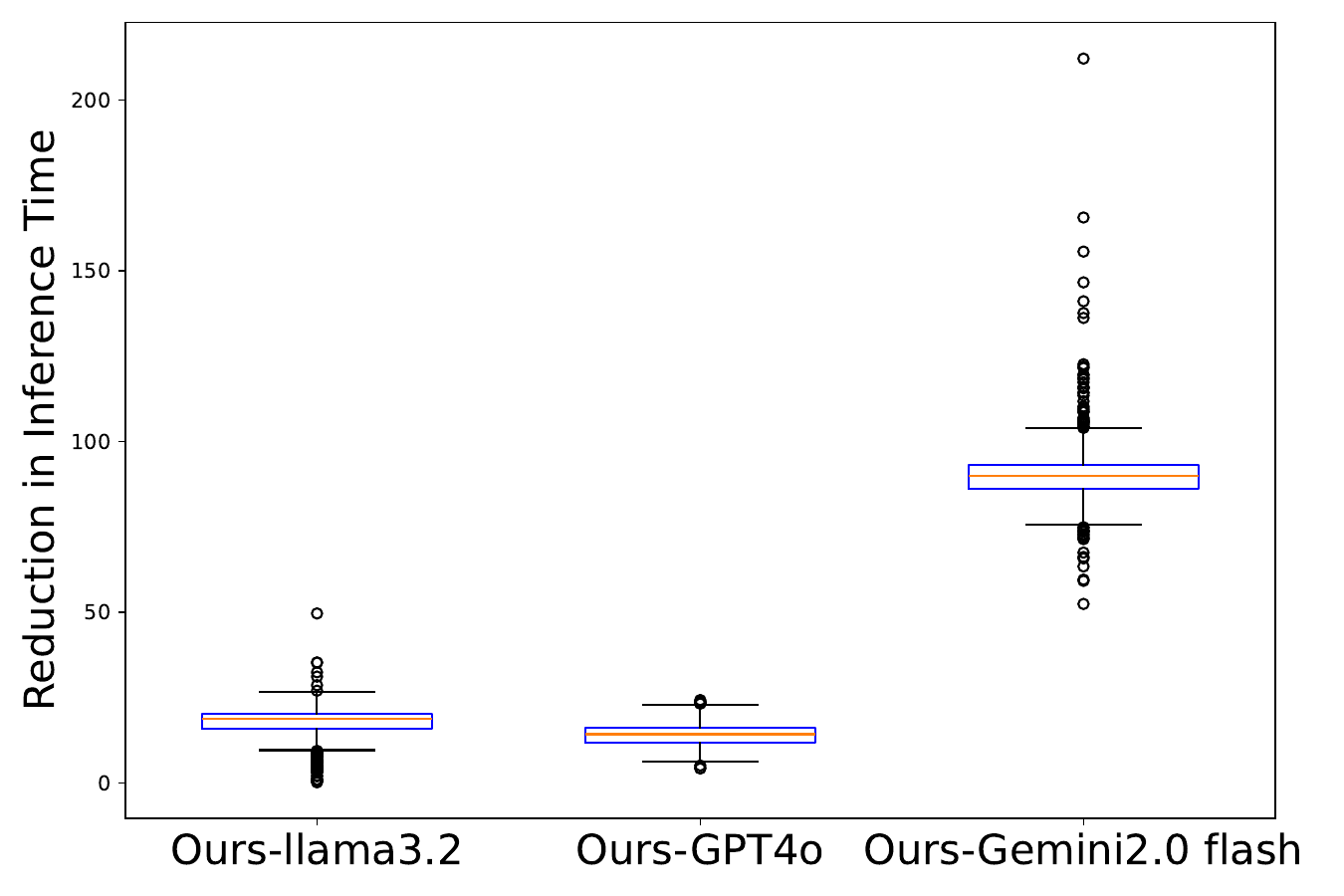}
    \caption{The inference time reduction of LLM-based methods over the naïve approach. Our Llama, GPT, and Gemini-based methods give a median reduction of 14X,18X, and 89X over naïve approach on all the datasets.}
    \label{fig:time}
  \end{subfigure}
  \caption{Comparison of training and inference time across different methods.}

\end{figure}

\textbf{Runtime Analysis.} The inference runtime statistics of our methods are presented in Table~\ref{time}, where our best Llama-based method achieves an inference time of 6.7 seconds for most time-series datasets. Additionally, as illustrated in Figure~\ref{fig:time}, LLM-based methods demonstrate a substantial reduction in inference time compared to the na\"ive approach, which involves evaluating all possible models and selecting the best-performing one. Specifically, our Llama, GPT, and Gemini-based methods achieve median inference time reductions of 14×, 18×, and 89×, respectively, over the na\"ive approach.

\begin{table}[ht]
    \centering
    \begin{tabular}{ccccc}
    \toprule
    \textbf{Methods}     &\textbf{Input Tokens} &\textbf{Output Tokens} \\
    \midrule

    \multicolumn{1}{l}{\method-Llama3.2}     &   \\
    \multicolumn{1}{r}{w. data} &$661.35\pm1.16$ &$157.14\pm508.51$  \\
    \multicolumn{1}{r}{w. data+CoT} &$739.35\pm1.16$ &$519.00\pm1445.28$ \\
    \multicolumn{1}{r}{w. data+meta\_features} &$20547.43\pm125.70$ &$116.75\pm274.24$\\
    \multicolumn{1}{r}{w. data+meta\_features+CoT} &$20632.43\pm125.70$ &$350.21\pm931.21$\\
    \midrule
    \multicolumn{1}{l}{\method-GPT4o}     &   \\
    \multicolumn{1}{r}{w. data}    &$2418.98\pm1472.00$ &$68.89\pm7.32 $ \\
    \multicolumn{1}{r}{w. data+CoT}     &$2711.98\pm1472.00$ &$297.18\pm46.68$\\
    \multicolumn{1}{r}{w. data+meta\_features} &$22475.06\pm1490.00$ &$67.54\pm8.37$  \\
    \multicolumn{1}{r}{w. data+meta\_features+CoT} &$22750.06\pm1490.00$ &$300.68\pm46.50$\\
    \midrule
    \multicolumn{1}{l}{\method-Gemini2.0 flash}     & \\
    \multicolumn{1}{r}{w. data} &$3075.64\pm2192.18$ &$84.33\pm7.28$\\
    \multicolumn{1}{r}{w. data+CoT} &$3075.64\pm2192.18$ &$340.53\pm60.29$\\
    \multicolumn{1}{r}{w. data+meta\_features} &$26761.32\pm2267.30$ &$80.99\pm4.06$\\
    \multicolumn{1}{r}{w. data+meta\_features+CoT} &$27080.32\pm2267.30$ &$352.31\pm80.54$\\
    \bottomrule

    \end{tabular}
    \caption{Input and output token count for each time series dataset.
    }
    \label{token}
\end{table}

\subsection{Ablation Studies and Additional Analyses}
\textbf{Meta-Features.} As shown in the Table [~\ref{tab:hitk},~\ref{time},~\ref{token}], incorporating meta-features in the prompt improves the performance of Llama and GPT-based methods, while the Gemini-based method appears to be less impacted. This improvement likely stems from the additional information provided by meta-features, which aids in selecting more suitable models. Besides, this performance gain comes at the cost of increased computational overhead— inference time rises by at least 25\%, and prompt token usage expands by at least 7X. 

\textbf{Chain-of-Thought Prompting.} Based on the Table [~\ref{tab:hitk},~\ref{time},~\ref{token}], explicitly incorporating CoT reasoning in the prompt—guiding the LLM to select the forecasting algorithm, hyper-parameters, and data representation step by step—does not necessarily enhance model selection performance and sometimes even degrades it while significantly increasing computational costs, leading to at least a 2X increase in inference time and a 4X rise in output token usage. We suspect that CoT prompting introduces unnecessary complexity, causing the LLM to overanalyze irrelevant aspects of the selection process. Unlike tasks where reasoning clarifies logic, model selection may benefit more from direct pattern recognition. The added reasoning steps could also increase the risk of hallucination, leading to suboptimal choices. 

\textbf{Data Representation.} Instead of allowing the LLM to select the data representation, we fixed it to either exponential smoothing or raw data. The results in Table~\ref{data} indicate that exponential smoothing enhances model selection performance for Llama and GPT-based methods, whereas it negatively impacts Gemini’s selection performance.

\begin{wrapfigure}{l}{0.4\textwidth} %
    \centering
    \includegraphics[width=0.4\textwidth]{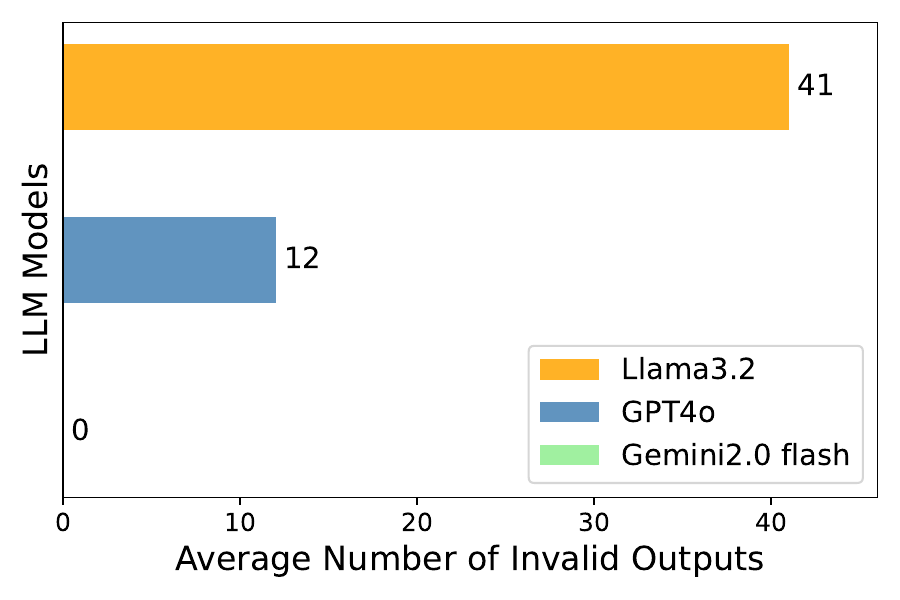}
    \caption{Average Number of Invalid Outputs for LLMs.}
    \label{fig:output}
\end{wrapfigure}

\textbf{Limitations of Different LLMs.} Llama3.2 achieves the best model selection performance among the three tested LLMs; however, it produces the most incomplete or irregular outputs as shown in Figure~\ref{fig:output}. In contrast, Gemini2.0 flash consistently generates complete and valid outputs while also having the lowest inference time and token usage. However, its performance is the weakest, under certain prompt settings, it even underperforms random selection. GPT4o serves as a balanced choice, delivering strong performance that surpasses almost all baselines, with only a few invalid outputs where the selected model falls outside the predefined model space.\\

\begin{table}[ht]
    \centering
    \resizebox{\columnwidth}{!}
    {
    \begin{tabular}{c|c|ccc}
    \toprule
    \multirow{2}{8em}{\textbf{hit@$k$ accuracy}} &\multirow{2}{4em}{\textbf{Methods}} &\multicolumn{3}{c}{\textbf{Data Representation}} \\ 
    
  & &LLM Selection &Exponential Smoothing & Raw \\
    \midrule
    \multirow{3}{4em}{hit@1}
    &\method-Llama3.2     &\textbf{1.14}&1.04 &0.31  \\
    &\method-GPT4o &0.52 &0.62 &0.31  \\
    &\method-Gemini2.0 flash &0.21&0 &0.42\\
    \midrule
    \multirow{3}{4em}{hit@5}
    &\method-Llama3.2     &4.47&\textbf{4.98} &3.32  \\
    &\method-GPT4o &2.60 &2.60&1.66  \\
    &\method-Gemini2.0 flash &0.62&0.31 &2.18\\
    \midrule
    \multirow{3}{4em}{hit@10}
    &\method-Llama3.2     &7.27 &\textbf{8.41} &4.47  \\
    &\method-GPT4o &4.88 &6.33&3.01  \\
    &\method-Gemini2.0 flash &3.53 &1.87 &2.70\\
    \midrule
    \multirow{3}{4em}{hit@50}
    &\method-Llama3.2     &29.60 &\textbf{31.15} &14.23  \\
    &\method-GPT4o &21.91 &23.36&11.84  \\
    &\method-Gemini2.0 flash &20.77 &19.94 &8.10\\
    \bottomrule

    \end{tabular}}
    \caption{hit@$k$ accuracy of LLM-based model selection, where the data representation is either chosen by the LLM or defaults to Exponential Smoothing or Raw.
    }
    \label{data}
\end{table}

\section{Conclusion, Limitations and Future Work}
In this work, we applied LLMs to the time-series forecasting model selection problem for the first time. Through extensive experiments, we demonstrated that LLMs can effectively address this task without relying on a precomputed performance matrix of historical model-dataset pair evaluations. Additionally, this method significantly reduces computational overhead, achieving up to 89X faster inference compared to exhaustive model evaluation. Despite the performance, the underlying mechanisms remain unclear. Furthermore, our current approach has been evaluated solely on univariate datasets. In future work, we aim to expand our testbed to incorporate a more diverse set of datasets and models, further exploring the generalizability of LLM-based model selection.

\section*{Ethics Statement}
Our research adheres to COLM Code of Ethics, ensuring that LLM-based model selection for time series forecasting is developed and applied responsibly. We prioritize fairness, transparency, and data privacy, avoiding biases that could impact decision-making across different forecasting applications. By leveraging LLMs for model selection without requiring an extensive historical performance matrix, our approach reduces potential biases introduced by past model rankings. Continuous ethical assessments guide our research to align with societal and regulatory standards.

\newpage
\bibliography{colm2025_conference}
\bibliographystyle{colm2025_conference}

\newpage
\appendix
\section{Appendix}
\subsection{Prompt Structure}
\label{appendix:a1}
The prompt structure we used is illustrated as follows.
\begin{tcolorbox}[width=\textwidth, sharp corners, boxrule=1pt, title={\textbf{Prompt and Response Structure}}]

\textbf{Prompt:} 
"[Role and Objective]...

 [Model Space]...

 [CoT Reasoning](optional)...
 
 [Input]...Dataset Values...Meta Features(optional)...

 [Output Format]...
 
 [Rules]..."

\textbf{Response:} 

\{"reasoning": "...", 

"result": \{
    "forecasting algorithm": "...",
    
    "hyperparameters": [
      \{"name": "...", "value": "..."\},
      \{"name": "...", "value": "..."\}
    ],
    
    "data representation": "..."
  \} \}
\end{tcolorbox}

We formulate the model selection problem following the framework of \citet{autoforecast}. 

\subsection{Datasets and Meta-Features}
\label{meta_feats}
Our approach relies on a collection of historical time-series forecasting datasets, denoted as ${\mathcal D} = \{D_1, D_2, \dots, D_N\}$, where $N$ is the total number of datasets. 
Each dataset \(D_i\) comprises a sequence of observations in \(\mathbb{R}^{n_i}\), with \(n_i\) representing the number of observations in \(D_i\).

For each dataset \(D_i \in \mathcal{D}\), we randomly sample \(T\) windows. Each time window \(w_t\) is a contiguous segment of observations from \(D_i\) with length \(|w_t|\) that is smaller than the total length of \(D_i\). For example, \(|w_{10}| = 16\) indicates that the 10th window contains 16 consecutive observations.

\textbf{Meta-Features Tensor.} To analyze the impact of meta-features on model selection in our approach, we utilize extracted meta-features for each time-series dataset from \citet{autoforecast}'s work.

\textbf{Definition 1.} Given a time-series dataset \(D_i\), we define the meta-features tensor \(\mathcal{F}_i = \{{F}_1^i, \dots, {F}_T^i\} \in \mathbb{R}^{T \times d}\), where the meta-features matrix \({F}_k^i \in \mathbb{R}^{d}\) captures the set of meta-features corresponding to the time window \(w_k\) of the dataset \(D_i\), given by
\[
{F}_k^i \triangleq \{\psi(w_k(D_i)) \mid \psi: \mathbb{R}^{|w_k|} \to \mathbb{R}^{d} \}, \tag{2}
\]
where \(\psi(\cdot): \mathbb{R}^{|w_k|} \to \mathbb{R}^{d}\) represents the feature extraction module and \(d\) denotes the number of the meta-features.\\
The extracted meta-features capture the key characteristics of each dataset and are grouped into five categories, as proposed by\citep{vanschoren2018metalearningsurvey}:
\begin{itemize}
    \item \textbf{Simple:} General task properties.
    \item \textbf{Statistical:} Properties of the underlying dataset distributions.
    \item \textbf{Information-theoretic:} Entropy measures.
    \item \textbf{Spectral:} Frequency domain properties.
    \item \textbf{Landmarker:} Forecasting models' attributes on the task.
\end{itemize}

\begin{table}[t]
    \centering
    \resizebox{\columnwidth}{!}
    {\begin{tabular}{l l l c}
        \toprule
        \textbf{Forecasting Algorithm} &\textbf{HyperParameter(s)} & \textbf{Data Representation} & \textbf{Total}\\
        \midrule
        DeepAR & num\_cells = [10,20,30,40,50] & \{Exp\_smoothing, Raw\} & 50 \\ 
        \citep{deepar} & num\_rnn\_layers = [1,2,3,4,5] & & \\
        \midrule
        DeepFactor & num\_hidden\_global = [10,20,30,40,50] & \{Exp\_smoothing, Raw\} & 50 \\
        \citep{deepar} & num\_global\_factors = [1,5,10,15,20] & & \\
        \midrule
        Prophet & changepoint\_prior\_scale = [0.001, 0.01, 0.1, 0.2, 0.5] & \{Exp\_smoothing, Raw\} & 50 \\ 
        \citep{prophet} & seasonality\_prior\_scale = [0.01, 0.1, 1.0, 5.0, 10.0] & & \\
        \midrule
        Seasonal Naive & season\_length = [1,5,7,10,30] & \{Exp\_smoothing, Raw\} & 10 \\
        \citep{seasonnaive} & & & \\
        \midrule
        Gaussian Process & cardinality = [2,4,6,8,10] & \{Exp\_smoothing, Raw\} & 50 \\ 
        \citep{gp} & max\_iter\_jitter = [5,10,15,20,25] & & \\
        \midrule
        Vector Auto Regression & cov\_type=\{``HC0'',``HC1'',``HC2'',``HC3'',``nonrobust''\} & \{Exp\_smoothing, Raw\} & 40 \\ 
        \citep{var} & trend = \{`n', `c', `t', `ct'\} & & \\
        \midrule
        Random Forest Regressor & n\_estimators = [10,50,100,250,500,1000] & \{Exp\_smoothing, Raw\} & 72 \\ 
        \citep{rf} & max\_depth = [2,5,10,25,50,'None'] & & \\
        \midrule
        & & & \textbf{322} \\
        \bottomrule
    \end{tabular}}
    \caption{Time-Series Forecasting Model Space. See hyperparameter definitions for various algorithms from GluonTS\citep{gluton} and statsmodels\citep{statmodel}. The number of models (last column) is all possible combinations of hyperparameters and data representations.}
    \label{modelspace}
\end{table}

\subsection{Performance Matrix}
\label{perfmat}
Now we introduce the performance matrix:

\textbf{Definition 2.} Given a training database \(\mathcal{D}\) and a model space \(\mathcal{M}\), we define the performance matrix \(\mathbf{P} \in \mathbb{R}^{T \times n \times m}\) as
\[
\mathbf{P} = \{\mathbf{P}_1, \mathbf{P}_2, \dots, \mathbf{P}_T\},
\]
where \(\mathbf{P}_k = (p_k^{i,j}) \in \mathbb{R}^{n \times m}\) and the element \(p_k^{i,j} = M_j(w_k(D_i))\) denotes the \(j\)th model \(M_j\)’s performance on the time window \(w_k\) of the \(i\)th training dataset \(D_i\). We denote
\[
p_k^i = \begin{bmatrix} p_k^{i,1} & \dots & p_k^{i,m} \end{bmatrix}
\]
as the performance vector of all models in \(\mathcal{M}\) on time window \(w_k\) of \(D_i\).

We denote the performance of a model on a time window using forecasting error metrics such as Mean Squared Error (e.g., MSE) of that model on that window.

\begin{table}
    \centering
    \begin{tabular}{ccccc}
    \toprule
    \textbf{Methods}     &\textbf{Training Time} &\textbf{Inference Time} \\
    \midrule
    \midrule
    Na\"ive    &N/A  &$70.9500 \pm 1.7801$  \\
    $ISAC^{*}$ & $278.8083 \pm 57.9900$   & $10.2480 \pm 2.7182$   \\
    $MLP^{*}$ &$705.2908 \pm 123.3715$  & $1.2745 \pm 0.5198$   \\
    \midrule
    \multicolumn{1}{l}{\method-Llama3.2}     &   \\
    \multicolumn{1}{r}{w. data} &N/A &$4.7201\pm16.9481$  \\
    \multicolumn{1}{r}{w. data+CoT} &N/A &$11.6174\pm32.7716$\\
    \multicolumn{1}{r}{w. data+meta\_features} &N/A &$6.7067\pm17.8676$ \\
    \multicolumn{1}{r}{w. data+meta\_features+CoT} &N/A &$17.1308\pm50.1011$\\
    \midrule
    \multicolumn{1}{l}{\method-GPT4o}     & &   \\
    \multicolumn{1}{r}{w. data}    &N/A &$1.4905\pm0.7234$ \\
    \multicolumn{1}{r}{w. data+CoT} &N/A &$4.7821\pm2.0229$\\
    \multicolumn{1}{r}{w. data+meta\_features} &N/A &$2.4630\pm0.7546$ \\
    \multicolumn{1}{r}{w. data+meta\_features+CoT} &N/A &$5.2368\pm1.4140$\\
    \midrule
    \multicolumn{1}{l}{\method-Gemini2.0 flash}     & & \\
    \multicolumn{1}{r}{w. data} &N/A &$0.7780\pm0.0543$ \\
    \multicolumn{1}{r}{w. data+CoT} &N/A &$2.1587\pm0.3407$\\
    \multicolumn{1}{r}{w. data+meta\_features} &N/A &$0.9790\pm0.0597$ \\
    \multicolumn{1}{r}{w. data+meta\_features+CoT} &N/A &$2.4974\pm0.4361$\\
    \bottomrule

    \end{tabular}
    \caption{Average and standard deviation inference and training runtime performance (in seconds) over all datasets.
    }
    \label{time}
\end{table}

\end{document}